\documentclass[sn-mathphys]{sn-jnl}

\usepackage{graphicx}%
\usepackage{multirow}%
\usepackage{amsmath,amssymb,amsfonts}%
\usepackage{amsthm}%
\usepackage{mathrsfs}%
\usepackage[title]{appendix}%
\usepackage{xcolor}%
\usepackage{textcomp}%
\usepackage{manyfoot}%
\usepackage{booktabs}%
\usepackage{algorithm}%
\usepackage{algorithmicx}%
\usepackage{algpseudocode}%
\usepackage{listings}%

\begin{document}

\title[Continual learning for surface defect segmentation by subnetwork creation and selection]{Continual learning for surface defect segmentation by subnetwork creation and selection}


\author[1]{\fnm{Aleksandr} \sur{Dekhovich}}

\author*[2]{\fnm{Miguel} A. \sur{Bessa}}\email{miguel\_bessa@brown.edu}

\affil[1]{\orgdiv{Department of Materials Science and Engineering}, \orgname{Delft University of Technology}, \orgaddress{\street{Mekelweg 2}, \city{Delft}, \postcode{2628 CD},  \country{The Netherlands}}}

\affil[2]{\orgdiv{School of Engineering}, \orgname{Brown University}, \orgaddress{\street{184 Hope St.}, \city{Providence}, \postcode{RI 02912}, \country{USA}}}


\abstract{

We introduce a new continual (or lifelong) learning algorithm called LDA-CP\&S that performs segmentation tasks without undergoing catastrophic forgetting. The method is applied to two different surface defect segmentation problems that are learned incrementally, i.e. providing data about one type of defect at a time, while still being capable of predicting every defect that was seen previously. Our method creates a defect-related subnetwork for each defect type via iterative pruning and trains a classifier based on linear discriminant analysis (LDA). At the inference stage, we first predict the defect type with LDA and then predict the surface defects using the selected subnetwork. We compare our method with other continual learning methods showing a significant improvement -- mean Intersection over Union better by a factor of two when compared to existing methods on both datasets. Importantly, our approach shows comparable results with joint training when all the training data (all defects) are seen simultaneously\footnote{Code implementation is available at: \url{https://github.com/adekhovich/continual_defect_segmentation}}.


}

\keywords{Continual learning, Automatic vision inspection, Surface defect segmentation, Linear discriminant analysis (LDA).}



\maketitle

\section{Introduction}\label{sec:introduction}

Automatic defects inspection plays an important role in product quality evaluation \citep{prunella2023deep}. In the beginning of the field, the creation of meaningful features to find defective regions was done manually \citep{ojala2002multiresolution, chao2008anisotropic, song2013noise, jeon2014defect}. Although classical machine learning methods have been proposed to identify images with defective surfaces \citep{jia2004intelligent, agarwal2011process, shanmugamani2015detection}, recent advances in deep learning research have led to an increase in performance \citep{prunella2023deep}. Typically, there are three types of tasks for defect inspection with neural networks -- classification, detection \citep{he2019end} and segmentation \citep{tabernik2020segmentation}. In the case of defect classification, transfer learning helps to increase the network's ability to detect defective surfaces \citep{aslam2020ensemble, wu2021hot}. For segmentation, most methods are based on the U-Net architecture \citep{ronneberger2015u} taking advantage of convolutional layers that automatically extract features from the images of the surfaces \citep{he2019defect, song2020edrnet, hao2021steel, huang2020surface}. Attention mechanisms \citep{vaswani2017attention} employed in the model's architecture can lead to even more accurate predictions \citep{pan2022dual, uzen2022swin}.

The advent of deep learning models came with more data for training and comparing these models in different real-life scenarios. For instance, after \cite{song2013noise} proposed their NEU-DET dataset with Hot Rolled Steel Strip Surface defects, containing six types of defects, other groups collected datasets with either different defect categories or a more significant number of defects, e.g., GC10-DET \citep{lv2020deep} and X-SDD \citep{feng2021x}. In segmentation literature, we can also find examples of different categorizations of surface defects, e.g., the Magnetic tile dataset \citep{huang2020surface} contains images of five types of defects together with defect-free cases. As a final example, the dataset collected by \cite{liu2022semi} also contains a large number of images but only has three types of defects.

Notwithstanding the increase in availability of datasets, there are many instances where there are few types of defects in each dataset. This is a natural occurrence in Engineering practice because many processes are not amenable to high-throughput. Simultaneously, if new defects occur or if another defect identification task with similar characteristics is encountered, using the original dataset and neural network model while considering new types of defects in similar (or even different) materials can be invaluable. However, training the same neural network model on a new dataset currently requires retraining it on all the data, even if the model was already capable of detecting some types of defects. This happens because deep learning models suffer from catastrophic forgetting \citep{french1999catastrophic,goodfellow2013empirical,coop2013ensemble}. In conventional training, neural networks cannot learn new tasks without forgetting old ones if the tasks are learned incrementally. Instead, the continual learning field \citep{de2021continual} aims to solve this type of problem where the model receives data in batches (tasks) but aims to learn information mitigating the forgetting issues.

\begin{figure*}[ht]
    \centering
    \includegraphics[width=\textwidth]{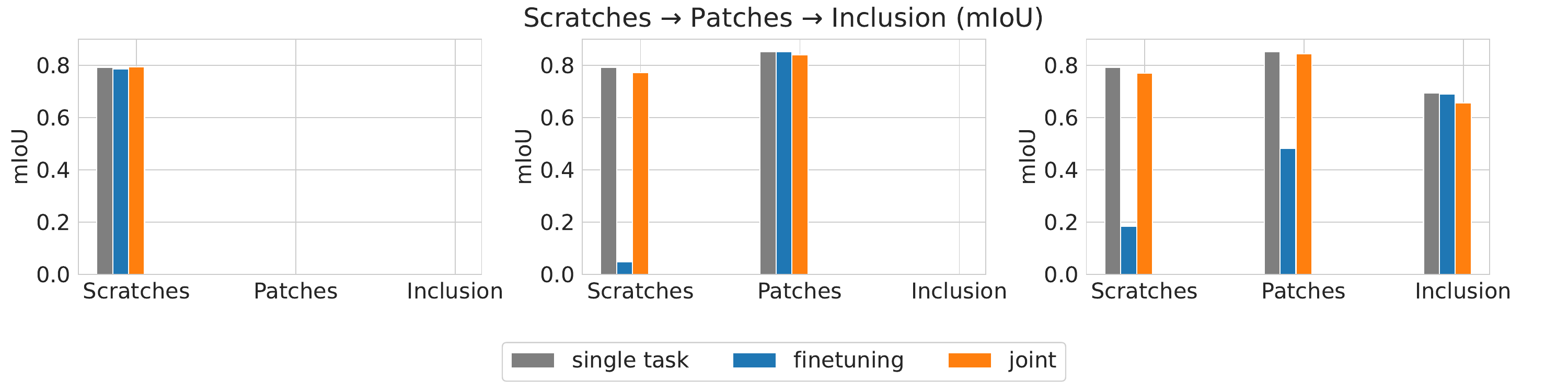}
    \caption{Example of forgetting in the case of incremental learning of three types of defects.}
    \label{fig:forgetting_example}
\end{figure*}

We illustrate the impact of catastrophic forgetting on segmentation tasks in Figure \ref{fig:forgetting_example} by considering the defect segmentation dataset SD-saliency-900 \citep{song2020saliency}. This dataset consists of images with three types of defects: scratches, patches and an inclusion. We illustrate this phenomenon by focusing on three typical learning scenarios: 1) \textit{single-task} training where each defect is learned with a single network, meaning there are three networks in total; 2) \textit{joint} training where the model has access to the entire dataset at once; 3) \textit{finetuning}, in which the network learns to segment sequentially, adapting the parameters for the new task, having them pretrained on previous ones.

For all three learning scenarios, we quantify the segmentation performance via the mean Intersection over Union (mIoU) score for every task after each incremental step (Figure \ref{fig:forgetting_example}). We observe that finetuning on a new task leads to a significant drop in performance for the previous task(s), as indicated by the blue bars -- a clear illustration that learning a sequence of tasks with a single network leads to forgetting the previous tasks in the sequence (catastrophic forgetting). However, forgetting does not occur in the case of single- and joint-task training because the network is capable of learning each of the defects separately without any pretraining, while also being capable of learning all of them together. We also note that both single- and joint-task training have comparable performance, despite a small decrease in the latter case\footnote{As a short note, marginal improvements in performance sometimes occur when changing the task order (investigated at the end of the article). For example, the mIoU performance for the Scratches task improved by 0.13 points after learning the Inclusion task, but the improvement is small compared to how much it degrades after learning the Patches task.}.

Therefore, we see that the ability to predict defects of previous types is lost when training for a new type of defect, i.e. that is out-of-distribution. The main objective of our work is to propose a continual learning algorithm suitable for the surface segmentation problem. To the best of our knowledge, this is the first work that develops a continual learning approach for surface defect segmentation.

\section{Continual learning}\label{sec:cl}

Overcoming the above-mentioned catastrophic forgetting requires deep learning models to be trained in a continual (or lifelong) learning manner \citep{thrun1998lifelong}. The overwhelming majority of continual learning literature is dedicated to classification tasks. In that context, three different categories have emerged \citep{de2021continual}: regularization-based \citep{li2017learning, zenke2017continual, aljundi2018memory}, replay-based \citep{rebuffi2017icarl,castro2018end, he2019end, douillard2020podnet} and architectural-based methods \citep{mallya2018packnet, sokar2022avoiding, dekhovich2023continual}. In some cases, there are methods that can fall into more than one category \citep{yan2021dynamically, wang2022foster}. Often these methods show better performance but require more memory or have high computational cost (e.g., extra memory buffer or architecture extension), which creates specific challenges when deployed in real-life applications. 

Regularization-based methods penalize parameters obtained on incremental step $t-1$ from drastic changes while learning the task on incremental step $t$. For example, SI \citep{zenke2017continual}, EWC \citep{kirkpatrick2017overcoming} and MAS \citep{aljundi2018memory} employ total loss $\mathcal{L}^{(t)}(x; \theta^{(t)})$ on incremental step $t$ that consists of the loss computed for the current data, and a penalty term to prevent forgetting:

\begin{equation}\label{eq:reg_loss}
    \mathcal{L}^{(t)} \big(x; \theta^{(t)} \big) = \mathcal{L}_{curr} \big(x; \theta^{(t)} \big) +  \frac{\lambda}{2} \sum_{i=1}^{\#params} \Omega_i \big(\theta_i^{(t)} - \theta_i^{(t-1)}\big)^2,
\end{equation}

\noindent where $\mathcal{L}_{curr}\big(x; \theta^{(t)}\big)$ is a loss on the current data, $\sum_{i=1}^{\#params} \Omega_i \big(\theta_i^{(t)} - \theta_i^{(t-1)}\big)^2$ is the penalty term, $\Omega_i$ is the cumulative importance for parameter $i$, and $\theta^{(t-1)}, \theta^{(t)}$ are network parameters at incremental steps $t-1$ and $t$, respectively. Learning without forgetting (LwF) \citep{li2017learning} aims to mitigate forgetting by minimizing the cross-entropy between output probabilities before and after the model is trained on a new task.

Replay-based (or rehearsal-based) approaches use a small fraction of data from previous tasks and keep it in a fixed-size memory buffer. However, storing old data in the buffer may not be allowed due to privacy issues \citep{zhang2020class}. Also, if the model parameters were downloaded without the memory buffer, further model training is not possible without forgetting. Therefore, in this work, we do not focus on this type of methods.

Architectural approaches do manipulations with the network structure by freezing and assigning some parameters to a specific task (e.g., PackNet \citep{mallya2018packnet}) or constantly growing the architecture increasing the expressivity of the network (e.g., DEN \citep{yoon2017lifelong}). However, if the model grows while learning a new task, the final number of parameters is not bounded, leading to additional computational costs. Alternatively, if the algorithm finds task-specific parameters, e.g., via iterative pruning in CP\&S \citep{dekhovich2023continual} or pruning at initialization in SupSup \citep{wortsman2020supermasks}, the challenge lies on activating the correct subnetwork during inference. This subnetwork selection in both CP\&S and SupSup requires a batch of test data of the task to be predicted, such that the correct subnetwork (i.e. task ID) is identified. This may also be impractical in real-life cases.

Literature on continual learning for semantic segmentation is scarcer. We can find examples that adapt classification continual learning algorithms to segmentation \citep{baweja2018towards, van2019towards}, or some new approaches designed specifically for segmentation \citep{klingner2020class, douillard2021plop, yan2021framework}. Similar to the classification case, better results are achieved by the methods that use a fixed-size memory buffer with samples from old tasks to overcome forgetting \citep{cha2021ssul, qiu2023sats}. However, even though these methods use old data (facilitating training), they still show significant forgetting of the first tasks while performing well only on the last ones.

Continual learning also finds its application in industrial and manufacturing cases. For example, MAS \citep{aljundi2018memory} was applied for product quality evaluation \citep{tercan2022continual}. The approach clones the output head for previous tasks with the lowest loss on the current data and uses this copy as initialization for a new task. The weight transfer for the output layer, and MAS algorithm that penalizes parameters from previous layers, show good performance for the considered regression problem. Regularization-based methods have been examined for anomaly detection in manufacturing process \citep{maschler2021regularization} and fault prediction
in lithium-ion batteries \citep{maschler2022regularization}. \cite{sun2022continual} developed an adaptive classification framework based on continual learning to identify new unlabeled samples. The proposed approach uses Mahalanobis distance and is employed to decide whether a new batch of data belongs to the already seen defect type, or forms a new one.

\section{Proposed approach}\label{sec:prop_approach}

We propose to take advantage of architectural methods that create task-specific subnetworks for each task, eliminating the subnetwork selection issue. As a base method, we consider Continual Prune-and-Select (CP\&S) \citep{dekhovich2023continual} where we improve the subnetwork selection process by training a model for this purpose, instead of having simple metric-based decision rules. In general, the task-prediction problem is quite challenging in continual learning \citep{kim2020class} and can be seen as an out-of-distribution (OOD) detection problem \citep{kim2022theoretical}. The difficulty arises from the presence of arbitrary classes in each task, leading to cases where classes within each task may not be similar, while classes from different tasks may have important similarities. This poses a challenge to identify the task-ID and corresponding subnetwork, affecting the performance of the continual learning model when the wrong subnetwork is selected. Conversely, these methods have the advantage that when the correct subnetwork is identified then there is no forgetting, which explains their state-of-the-art performance in different image-classification datasets \citep{dekhovich2023continual}.

However, in contrast to image classification, every task in defect segmentation problems consists of defects of only one type. This represents an opportunity for architectural continual learning methods because we can train a model that learns the distribution of each defect separately. To do so, we use linear discriminant analysis trained on features extracted from a pretrained convolutional neural network \citep{dorfer2015deep, hayes2020lifelong}. For the segmentation model, we use the U-Net architecture \citep{ronneberger2015u}, in which we create task-specific subnetworks via iterative pruning. As a pretrained feature extractor, we use the EfficientNet-B5 architecture \citep{tan2019efficientnet} pretrained on ImageNet-1000 \citep{deng2009imagenet}.

\begin{figure*}
    \centering
    \includegraphics[width=\textwidth]{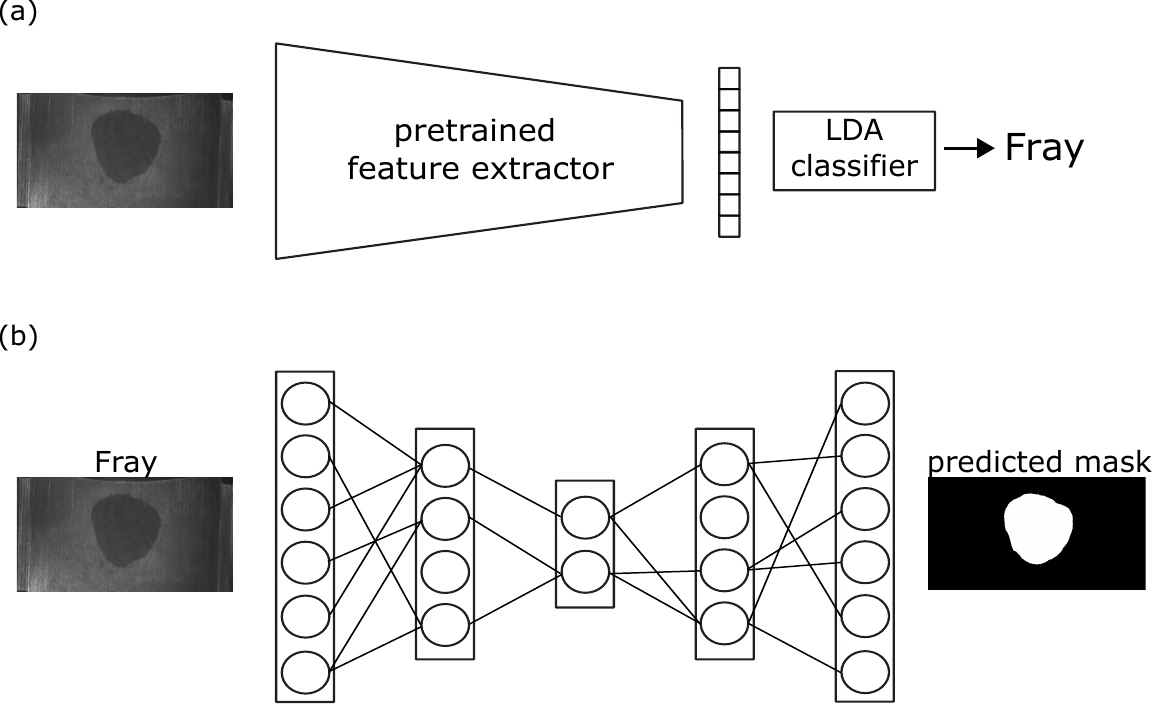}
    \caption{An overview of the proposed method: (a) task ID (i.e., defect type/subnetwork) prediction; (b) defects segmentation.}
    \label{fig:lda-cps}
\end{figure*}

In Figure \ref{fig:lda-cps}, we illustrate the inference stage of our approach, which consists of two steps: (1) predicting the defect type (task ID) with LDA; and (2) using a subnetwork that corresponds to the predicted defect to predict the segmentation mask. Note that at the inference stage, defect type prediction and defect mask prediction need to be done sequentially. Training for these steps can be done in parallel and independently from each other. We call our proposed approach LDA-CP\&S since it uses the CP\&S paradigm of creating subnetworks during training, and it employs LDA for the subnetwork selection.

Referring again to Figure \ref{fig:forgetting_example}, we recall that the three separate models (single-task grey bars) are capable of learning the defects slightly better than joint training with all the tasks together (orange bars). This hints that having task-specific parameters associated with only one task can even help the learning process. At the same time, the shared parameters provide a transfer learning effect between a new subnetwork and all the ones created before. Both advantages can be exploited by LDA-CP\&S.

Notwithstanding, our method could suffer a performance drop from two possible sources: the pruning stage, and the LDA classification stage. The performance reduction due to pruning may occur because some important parameters could be deleted when creating additional space (free connections) for future tasks. In addition, misclassification by LDA could result in signal routing through the wrong subnetwork and consequently poor segmentation performance. In Section \ref{sec:results}, we show that these two sources of error are negligible compared to the benefits of our approach. In the following subsections, we describe the processes for subnetwork creation and LDA training. 

\subsection{Subnetwork creation}

To create a subnetwork for the given task, we use NNrelief pruning algorithm \citep{dekhovich2021neural}. The approach evaluates the strength of the signal that propagates through every connection/kernel. This pruning technique shows better sparsity results than other connection/kernel-based pruning techniques \citep{han2015learning, li2016pruning, lee2018snip}.

For the set of $m_{l-1}$-channelled input samples $\mathbf{X}^{l-1} = \{\mathbf{x}^{l-1}_1, \ldots, \mathbf{x}^{l-1}_N\}$, where  $\mathbf{x}^{(l-1)}_k = (x^{l-1}_{k1}, \ldots, x^{l-1}_{km_{l-1}}) \in \mathbb{R}^{m_{l-1} \times h_{l-1}^1 \times h^2_{l-1}}$ with $h^1_{l-1}$ and $h^2_{l-1}$ being the height and width of feature maps for convolutional layer $l$. For every kernel $\mathbf{K}^{l}_{1j}, \mathbf{K}^{l}_{2j}, \ldots, \mathbf{K}^{l}_{m_l j}, \ \mathbf{K}^{l}_{ij} = (k^{l}_{ijqt}) \in \mathbb{R}^{r_l \times r_l}$, $q \geq 1, \ r_l \geq t$, where $r_l$ is a kernel size, and for every bias $b^{(l)}_j$ in filter $\mathbf{F}^{l}_j$, we define $\mathbf{\hat{K}}^{l}_{ij} = \Big( \Big\lvert k^{l}_{ijqt} \Big\rvert \Big)$ as a matrix consisting of the absolute values of the matrix $\mathbf{K}^{(l)}_{ij}$. Then we compute importance scores $s^{l}_{ij}, i \in \{1, 2, \ldots, m_l\}$ of kernels $\mathbf{K}^{l}_{ij}$ as follows:

\begin{align}
    \label{eq:conv_importance_scores}
    s^{l}_{ij} &= \frac{\frac{1}{N}\sum_{n=1}^N \Big\lvert\Big\lvert \mathbf{\hat{K}}^{l}_{ij} * \Big\lvert x^{l-1}_{ni} \Big\rvert \ \Big\rvert\Big\rvert_F}{S^{l}_{j}},
\end{align}

\noindent where $S^{l}_{j} = \sum_{i=1}^{m_{l-1}}\Big(\frac{1}{N}\sum_{n=1}^N \Big\lvert\Big\lvert \mathbf{\hat{K}}^{l}_{ij}* \Big\lvert x^{l-1}_{ni} \Big\rvert \ \Big\rvert\Big\rvert_F\Big)$ is the total importance score in filter $\mathbf{F}^{l}_j$ of layer $l$, with $*$ indicating a convolution operation, and where $\lvert \lvert\cdot \rvert \rvert_F$ is the Frobenius norm. 

The sketch of the algorithm for pruning filter $\mathbf{F}^{l}_j$ in a convolutional layer $l$ can be described as follows:
\begin{enumerate}
    \item Choose $\alpha \in (0, 1)$ -- the amount of kernels' importance that we want to keep relative to the total importance of the kernels in the filter $\mathbf{F}^{l}_j$.
    \item Compute importance scores $s_{ij}^{l}$ for all kernels in the filter $\mathbf{F}^{l}_j, \ i=1,\ldots, m_{l-1}$, using Eq. \ref{eq:conv_importance_scores}. 
    \item Sort importance scores $s_{ij}^{l}$ for the filter $\mathbf{F}^{l}_j$.
    \item For the sorted importance scores $\hat{s}_{ij}^{l}$ find minimal $p \le m_{l-1}$ such that $\sum_{i=1}^p \hat{s}_{ij}^{l} \ge \alpha$.
    \item Prune kernels with the importance score $s_{ij}^{l} < \hat{s}_{pj}^{l}$ for all $i \le m_{l-1}$ and fixed $j$. 
\end{enumerate}

\begin{figure*}
    \centering
    \includegraphics[width=\textwidth]{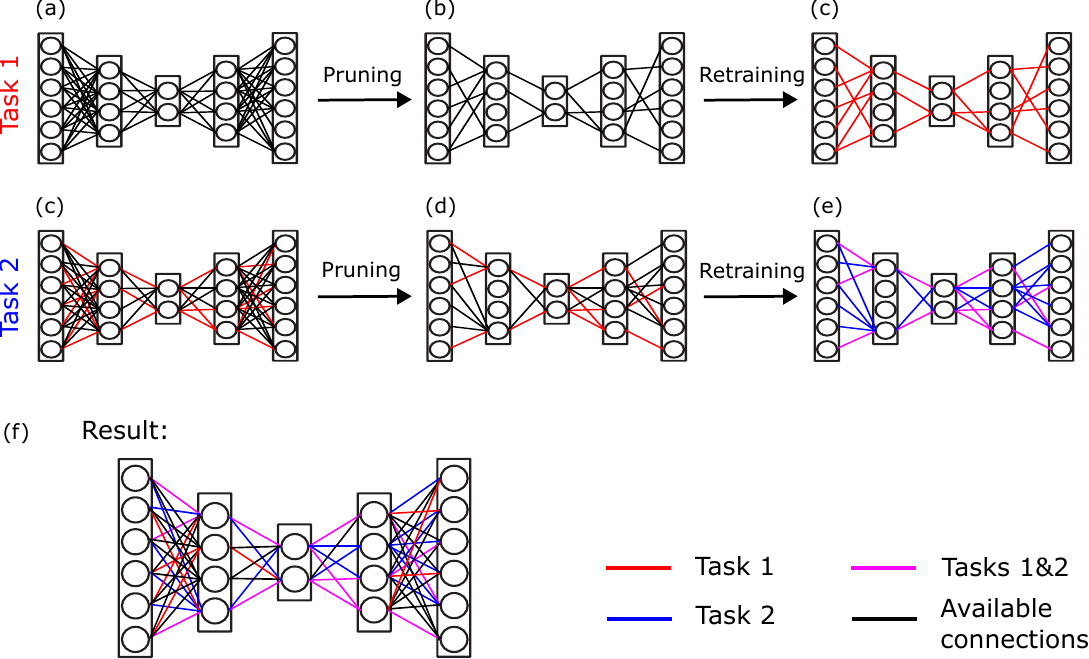}
    \caption{An overview of Continual Prune-and-Select (CP\&S): an example with two tasks.}
    \label{fig:cps}
\end{figure*}

Overall, NNrelief finds kernels that propagate on average the lowest signal according to the Frobenious norm and prune these kernels. As the outcome of the procedure, we obtain a subnetwork (sub-U-Net) that predicts the defect for only one type of defects. Then we fix all parameters that are assigned to this subnetwork and do not update them anymore. When the network receives a new task with a new type of defect, CP\&S finds a subnetwork for this task within the main U-Net, using the parameters assigned to the previous tasks, but without updating them. Algorithm \ref{alg:cps_training} illustrates the pseudocode for CP\&S and Figure \ref{fig:cps} illustrated the method:

\begin{algorithm}[H]
    \begin{algorithmic}[1]
        \Require network $\mathcal{N}$, dataset $\{\textbf{X}^t\}_{t=1}^T$.
        Initialize {learning parameters (learning rate, weight decay, number of epochs, etc. ), pruning parameters (for NNrelief algorithm: $\alpha$ and $num\_iters$).}
        \For {$t=1, 2, \ldots, T$}
            \State{$\mathcal{N}^{(t)} \gets \mathcal{N}$}
            \For {$it = 1,2, \ldots, num\_iters$}    \Comment{repeat pruning}
                \State{$\mathcal{N}^{(t)} \gets \text{Pruning}(\mathcal{N}^{(t)}, \mathbf{X}^{(t)}, \alpha)$} \Comment{pruning step: NNrelief}
                \State{Retrain subnetworks $\mathcal{N}^{(t)}$}     \Comment{retraining step}
            \EndFor
            \State{Freeze parameters $w \in \mathcal{N}^{(t)}$ and never update them}
        \EndFor
        \Ensure network $\mathcal{N}$ that learned tasks $1, 2, \ldots, T$.
    \end{algorithmic}
    \caption{Pseudocode for CP\&S training procedure}
    \label{alg:cps_training}
\end{algorithm}

\subsection{Subnetwork prediction (or selection)}

To predict the task ID (type of defect) at the inference stage, we propose to use linear discriminant analysis (LDA). In this subsection, we describe the training procedure for LDA. In LDA, it is assumed that all classes have class means $\mathbf{\mu}^{(1)}, \mathbf{\mu}^{(2)}, \ldots, \mathbf{\mu}^{(T)}$ and share the same covariance matrix $\mathbf{\Sigma}$. However, in continual learning, we do not have access to all tasks at the same time, but only task $t$. Therefore, the covariance matrix needs to be updated online with respect to the new data batch.

Let us denote a new given task as $\mathbf{X}^{(t)} = \{x_1^{(t)}, x_2^{(t)}, \dots, x_{N_t}^{(t)}\}$. Following streaming LDA (SLDA) strategy \citep{hayes2020lifelong}, we use a feature extractor $\mathcal{F}$ pretrained on ImageNet-1000 to obtain low-dimensional data representation $\mathbf{Z}^{(t)} := \{z^{(t)}_1, z^{(t)}_2, \ldots, z^{(t)}_{N_t}\}$,  $z^{(t)}_i = \mathcal{F}(x_i^{(t)}) \in \mathbb{R}^{d}$. Then we can compute the class mean $\mathbf{\mu}^{(t)} \in \mathbb{R}^{d}$ and update the shared covariance matrix $\mathbf{\Sigma}^{(1:t)} \in \mathbb{R}^{d \times d}$ after incremental step $t$ as follows \citep{dasgupta2007line}:

\begin{align}
    \mathbf{\mu}^{(t)} &= \frac{1}{N_t}\sum_{i=1}^{N_t} z_i^{(t)}\\
    \mathbf{\Sigma}^{(1:t)} &= \frac{(t-1) \mathbf{\Sigma}^{(1:t-1)} + \mathbf{\Delta}^{(t)}}{t},
\end{align}
where $\mathbf{\Delta}^{(t)} = \frac{(t-1) (Z^{(t)} - \mu^{(t)})(Z^{(t)} - \mu^{(t)})^\mathsf{T}}{t}$ and $(Z^{(t)} - \mu^{(t)}) := (z_1^{(t)} - \mu^{(t)}, z_2^{(t)} - \mu^{(t)}, \ldots, z_{N_t}^{(t)} - \mu^{(t)}) \in \mathbb{R}^{d \times N_t}$. In SLDA, the regularized version of LDA is implemented by applying shrinkage regularization to covariance matrix: $\mathbf{\Lambda}^{(1:t)} = [(1-\varepsilon)\mathbf{\Sigma}^{(1:t)} + \varepsilon \mathbf{I}]^{-1}$, where $\mathbf{I}$ is an identity matrix of the corresponding dimension.

At the inference stage, after learning all class means $\mathbf{\mu}^{(t)}, \ t=1,2,\ldots, T$, and shared covariance matrix $\mathbf{\Sigma}^{(1:T)}$ (and $\mathbf{\Lambda}^{(1:t)}$ as a result), we can make a prediction for a new test sample $x$ as follows:

\begin{equation}
    c = \underset{i=1, 2, \ldots, T}{\text{argmax }} (\mathbf{W} \mathcal{F}(x) + \mathbf{b})_i,
\end{equation}

\noindent where $\mathbf{W} = \mathbf{M}^{(1:T)} \mathbf{\Lambda}^{(1:T)}$, rows of $\mathbf{M}^{(1:T)}$ are mean vectors $\mathbf{\mu}^{(t)} \ (t=1, 2, \ldots, T)$, and $\mathbf{b}_i = -\frac{1}{2} \mu^{(i)} \mathbf{\Lambda}^{(1:T)} \mu^{(i)}$.

Unlike previous task prediction strategies \citep{wortsman2020supermasks, rajasegaran2020itaml}, with LDA we can predict the task ID with a single test sample, rather than with a batch of samples, representing an important advantage. This is possible because each task consists of defects of the same type and can be described well by a normal distribution with class means $\mathbf{\mu}^{(t)}$ and common covariance matrix $\mathbf{\Sigma}^{(1:T)}$.

\section{Experiments and results}\label{sec:results}

We evaluate our LDA-CP\&S approach on the SD-saliency-900 \citep{song2020saliency} and Magnetic tile defects \citep{huang2020surface} datasets, comparing with the following scenarios:

\begin{itemize}
    \item joint training: the model has access to all data at each incremental step. This case is an upper bound for rehearsal-based methods.
    \item finetuning: the model is trained at each incremental step $t$ without preventing forgetting, i.e., we finetune the model to a new task $t$ that is pretrained on previous tasks $1, 2, \ldots, t-1$, inevitably causing forgetting of previous tasks because the network parameters (weights and biases) are updated for task $t$.
    \item Regularization-based continual learning methods: LwF, MAS that penalize important parameters from changing (see Section \ref{sec:cl}, Eq. \ref{eq:reg_loss}), in an attempt to alleviate forgetting.
\end{itemize}

We do not consider rehearsal-based approaches that replay a small portion of data from previous tasks while learning a new one because our premise is that old data is not available and should not be used. Furthermore, our comparative investigation of the proposed LDA-CP\&S method with others includes the joint training strategy, which is an upper bound for rehearsal-based methods, where all data is available at each incremental step. Therefore, if we show that LDA-CP\&S performs similarly to joint training, there is no need to consider rehearsal-based continual learning methods.

As performance metrics, we follow other segmentation works and use the mean Intersection over Union score. For ground truth $Y$ and prediction $\hat{Y}$, IoU score is computed as follows:
\begin{align}
    \text{IoU}(\hat{Y}, Y) = \frac{\lvert Y \cap \hat{Y} \rvert}{\lvert Y \cup \hat{Y}\rvert}.
\end{align}
To train the model, we use IoU loss which leads to better performance in our experiments than other losses, e.g., Tversky loss \citep{salehi2017tversky} and Focal loss \citep{lin2017focal}. However, it is worth noting that the difference in IoU scores between models trained with different loss functions is not significant. The IoU loss is computed as follows:
\begin{align}
    \text{IoUloss}(\hat{P}, Y) = 1 - \frac{\sum_{i=1}^H \sum_{j=1}^W \hat{P}_{ij} \cdot Y_{ij} + \varepsilon}{\sum_{i=1}^H \sum_{j=1}^W \hat{P}_{ij} + Y_{ij} - \hat{P}_{ij} \cdot Y_{ij} + \varepsilon},
\end{align}

\noindent where $\hat{P}_{ij} \in [0, 1]^{H \times W}$ are the output probabilities, and $H$ and $W$ are the height and width of the output image, $\varepsilon$ is a smoothing parameter.   

\subsection{SD-saliency-900 dataset}

In the case of the SD-saliency-900 dataset, we consider a smaller version of U-Net with 16, 32, 64 and 128 in the encoder block and 256 channels in the bottleneck because it consists of only three types of defects -- Scratches, Patches and Inclusion -- with 300 images per defect. The original size of the images is $200 \times 200$ but we resize the images to $224 \times 224$ to make them acceptable for U-Net. We train the segmentation model for 70 epochs with 8 images in a batch, using Adam \citep{kingma2014adam} optimizer and learning rate $0.001$. During the pruning stage, we use $\alpha = 0.9$ and $3$ pruning iterations. More details about the influence of hyperparameters on the results are shown in Section \ref{sec:hyperparameters}. As it is common in continual learning literature \citep{masana2020class}, we consider different task orderings in our experiments. We can construct six task orderings for the current dataset (e.g., Patches $\to$ Scratches $\to$ Inclusion). 

\begin{table*}[ht]
    \begin{center}
        \caption{Classification accuracy (\%) for SD-saliency-900 dataset. The numbers are averaged over all six orderings.}
        \begin{tabular}{ @{} lllllc @{} }
            \toprule
              & Scratches & Patches & Inclusion & & Average\\
            \midrule
            accuracy (\%) & 98.33 & 100 & 100 &  & 99.44\\
            \botrule
        \end{tabular}
        \label{tab:sd900-class_acc}
    \end{center}
\end{table*}

First, we have to make sure that LDA can accurately predict the defect type in an incremental manner. Table \ref{tab:sd900-class_acc} illustrates the classifier's accuracy for each defect averaged over all six orders. We can observe the high performance of LDA, misclassifying only a few images from the Scratches dataset. Since 60 images were selected to test each defect type, the prediction error presented corresponds to only 1 misclassified image.

\begin{figure*}[ht!]
    \centering
    \includegraphics[width=\textwidth]{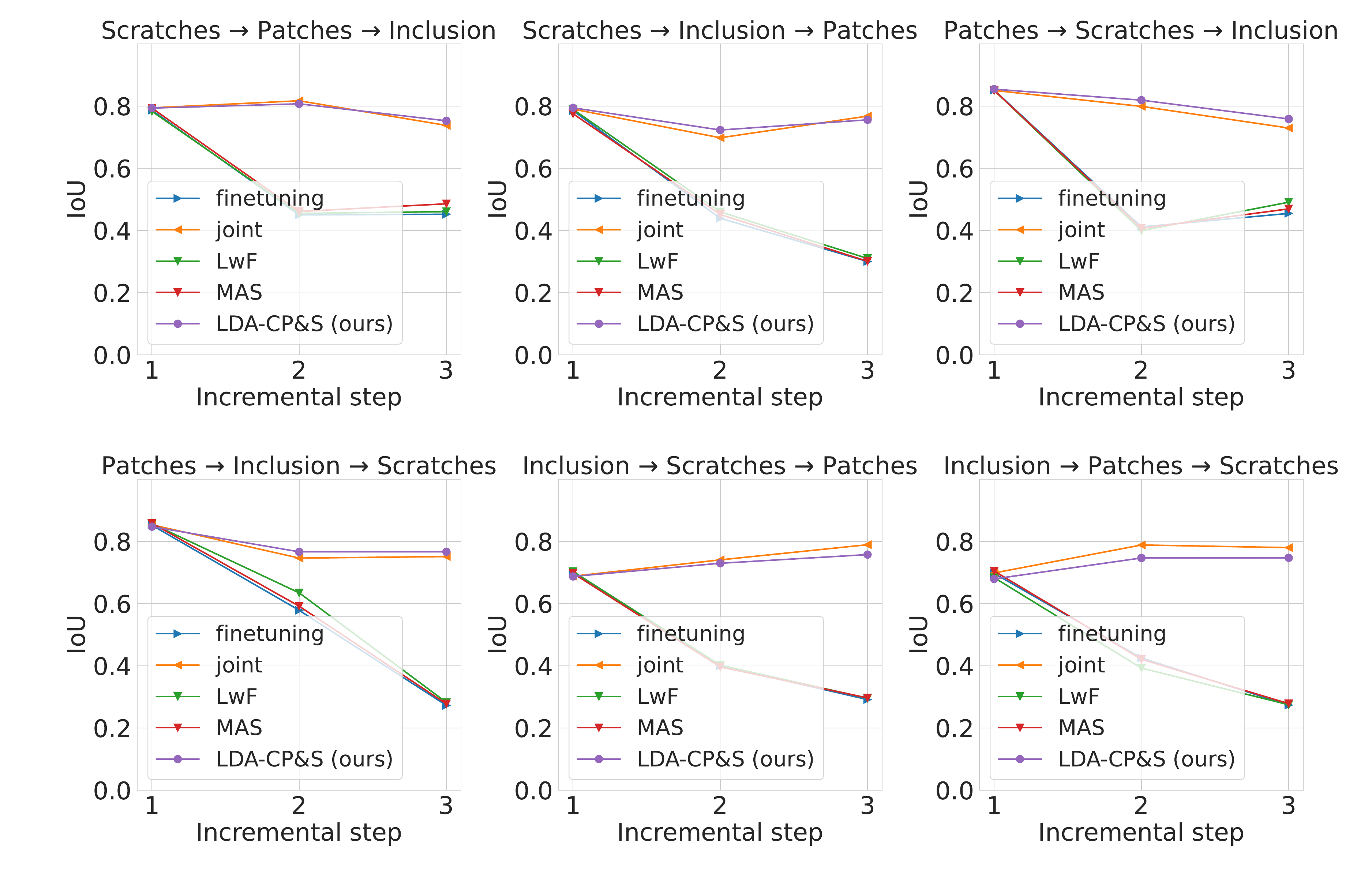}
    \caption{IoU score after every incremental step for SD-saliency-900 dataset. The results are presented for all six possible defect orderings.}
    \label{sd-900_iou_6orders_avg}
\end{figure*}

In Figure \ref{sd-900_iou_6orders_avg}, we show mIoU score after every incremental step for every task order. Regularization-based methods only slightly outperform finetuning strategy, while our LDA-CP\&S shows comparable results to joint training. Poor performance of the regularization methods can be explained by the lack of a task-specific output layer, which is present in classification network architectures as a classification head. Therefore MAS and LwF update all parameters but change them slightly less than finetuning. On the contrary, LDA-CP\&S creates fixed task-specific subnetworks that can overlap and transfer knowledge between each other. Since LDA predicts the defect type (i.e., subnetwork) well at the inference stage, we almost do not have any losses in segmentation performance. We also do not observe network saturation, i.e., the situation when the model does not have enough free space to learn a new task, even though we use a smaller version of U-Net.

\subsection{Magnetic tile defects dataset}

Magnetic tile defects dataset \citep{huang2020surface} contains five types of defects, namely Blowhole, Break, Crack, Fray and Uneven, and images that are free from defects (Free). In this work, we consider only images with defects, i.e., five classes. Since the number of defects is higher in this case, we use a U-Net of the original size with 64, 128, 256, and 512 in the encoder block and 1024 channels in the bottleneck. All images in the dataset have different image sizes and, therefore, we resize them to $224 \times 224$. For every defect, we randomly select 80\% images for training and the rest for testing. U-Net is trained for 150 epochs with 8 images in a batch, using Adam optimizer and learning rate $0.0001$. Since the total number of possible task orderings is quite large ($5! = 120$), we consider only five of them at random and we do not have reason to believe that the final performance would be very different when choosing other orderings:
\begin{itemize}
    \item Blowhole $\to$ Break $\to$ Crack $\to$ Fray $\to$ Uneven;
    \item Break $\to$ Uneven $\to$ Fray $\to$ Crack $\to$ Blowhole;
    \item Crack $\to$ Blowhole $\to$ Break $\to$ Uneven $\to$ Fray;
    \item Fray $\to$ Crack $\to$ Uneven $\to$ Blowhole $\to$ Break;
    \item Uneven $\to$ Fray $\to$ Blowhole $\to$ Break $\to$ Crack,
\end{itemize}

\noindent where each defect type appears exactly once for each ordering.

One of the main difficulties with this dataset is class imbalance. Table \ref{tab:magnetic-tile_class_acc} presents LDA accuracy with the same feature extractor considered in the previous example: the pretrained EfficientNet-B5 architecture. Overall, our classification model is able to identify correctly four out of five types of defects, having some difficulties with the Fray sub-dataset that contains the smallest number of images. The only mistake was done in the Fray sub-dataset where we have only 7 test images, meaning that only one image is classified wrongly. 

\begin{table*}[ht]
    \begin{center}
        \caption{Classification accuracy (\%) for Magnetic tile dataset and the total size of the dataset. The numbers for accuracy are averaged over all five orderings.}
        \begin{tabular}{ @{} lllllllc @{} }
            \toprule
              & Blowhole & Break & Crack & Fray & Uneven & & Average\\
            \midrule
            \# train (test) images  & 92 (23) & 92 (22) & 68 (17) & 25 (7) & 72 (21) &  &  N/A\\  
            \midrule
            test accuracy (\%) & 100 & 100 & 100 & 85.71 & 100 & & 98.75\\
            \botrule
        \end{tabular}
        \label{tab:magnetic-tile_class_acc}
    \end{center}
\end{table*}

We would like to highlight the necessity of pretraining the network that extracts features for LDA. The pretrained EfficientNet-B5 produces lower dimensional embeddings that can be used for training and classification with an accuracy of 98.75\%, misclassifying only one test image. Meanwhile, if we were to consider a feature extractor with random parameters it would compress the input images in such a way that the LDA classifier would only achieve 16.24\% of accuracy.

\begin{figure*}[ht!]
    \centering
    \includegraphics[width=\textwidth]{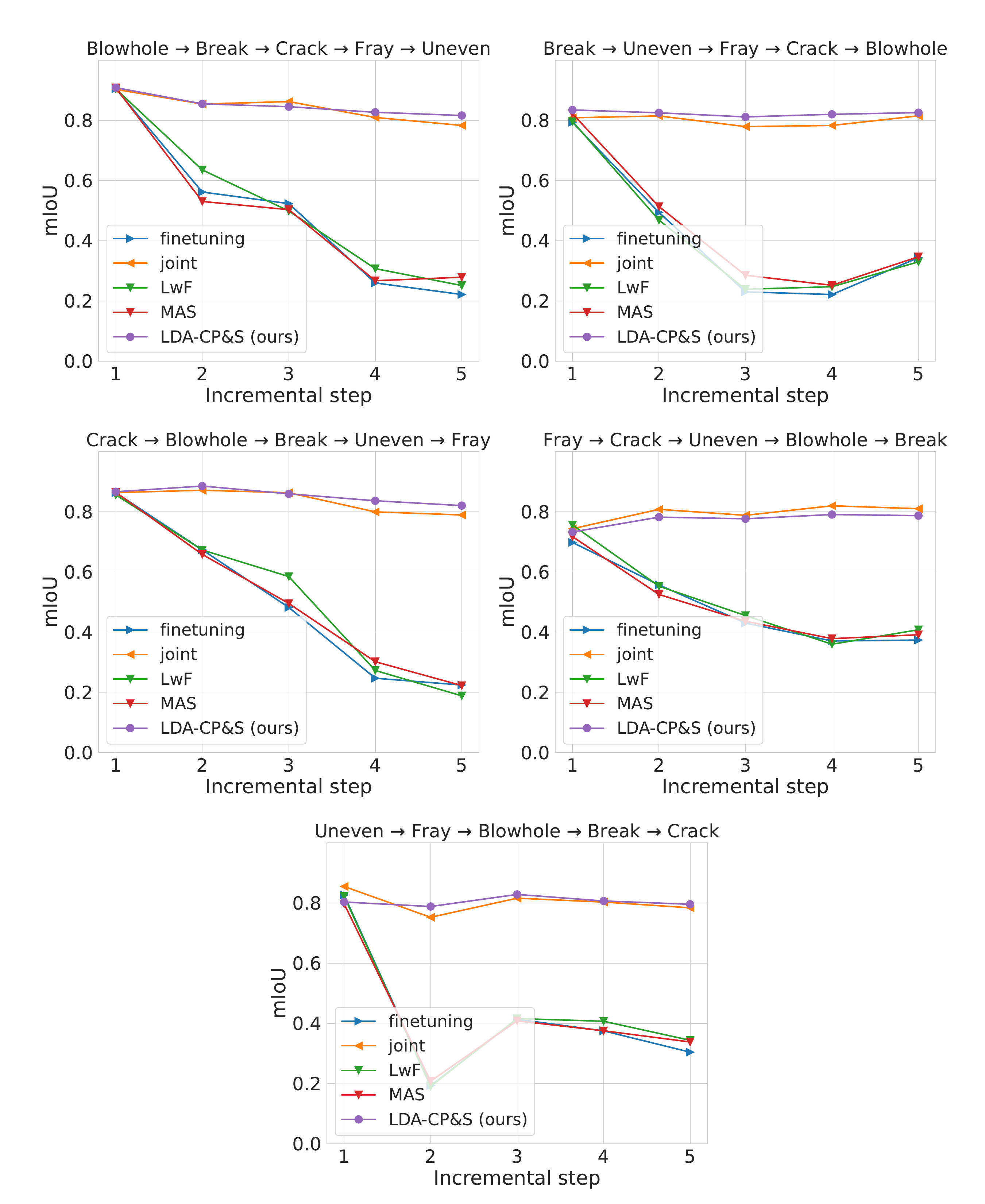}
    \caption{IoU score after every incremental step for Magnetic tile datasets. The results are shown for all five selected defect orderings.}
    \label{fig:magnetic-tile_iou_5orders}
\end{figure*}

In Figure \ref{fig:magnetic-tile_iou_5orders}, we present the mIoU score after every incremental step, comparing our LDA-CP\&S with other continual learning methods. As we saw in the previous example, regularization-based methods do not handle this type of segmentation problem well. The tasks that we constructed from the Magnetic tile dataset can be quite dissimilar having significant differences in defects areas. Therefore, by updating all the parameters without having task-specific ones, regularization-based approaches are only slightly better than simple finetuning where no anti-forgetting measures are considered. In contrast, our LDA-CP\&S creates task-specific parameters for each defect, fixing the values of the parameters once they are assigned to a subnetwork (i.e., defect type or task ID). This allows LDA-CP\&S to deal with sequences of tasks as well as joint training, which is very encouraging because joint training is a performance upper bound since all the data is available at each incremental step.

\begin{figure*}[ht!]
    \centering
    \includegraphics[width=\textwidth]{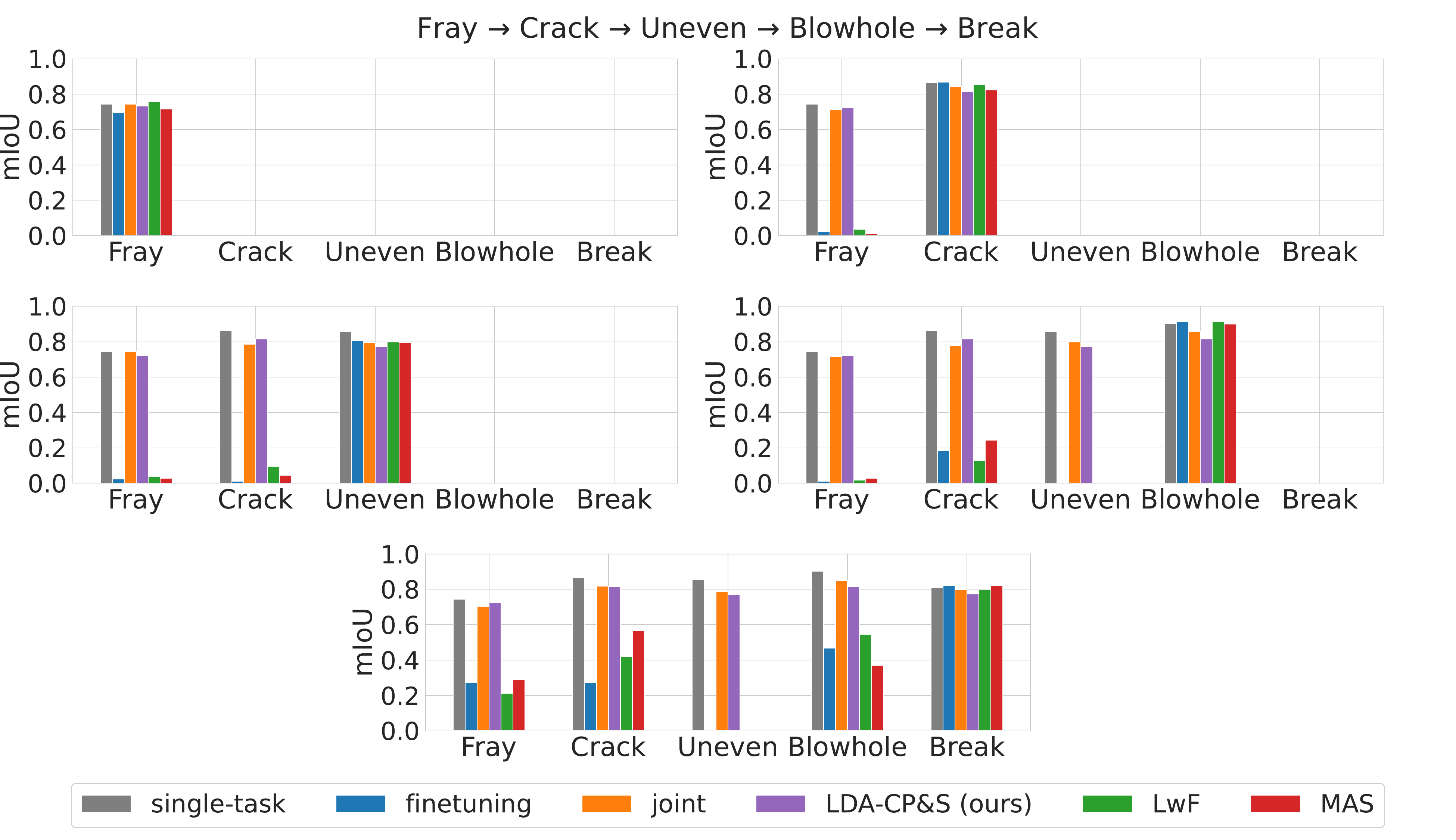}
    \caption{IoU score after every incremental step for one of the defects orderings from the Magnetic tile dataset.}
    \label{fig:magnetic-tile_order3_iou_by-task.pdf}
\end{figure*}

We also investigated how the mIoU score changes for every task after each incremental step. In Figure \ref{fig:magnetic-tile_order3_iou_by-task.pdf}, we consider one of the task orderings: Fray $\to$ Crack $\to$ Uneven $\to$ Blowhole $\to$ Break. The figure clearly shows the advantage of our algorithm over regularization-based ones because they are heavily dependent on the similarity of the tasks in the order. For example, learning the Break sub-dataset (the last incremental step) improves performance on Fray and Crack sub-datasets compared to the previous incremental step for MAS, LwF and finetuning strategies. However, Uneven is totally forgotten after the network is trained on the Blowhole sub-dataset.

On the contrary, LDA-CP\&S does not forget previous tasks and is still able to learn new ones even having fewer free parameters. It has comparable performance with a single-task scenario, where a separate U-Net is trained for every task. Also, we observe that task-wise performance is almost the same as for joint training, meaning the subnetwork overlaps provide enough knowledge transfer to learn a new task. 

\begin{figure}
    \centering
    \includegraphics[width=\textwidth]{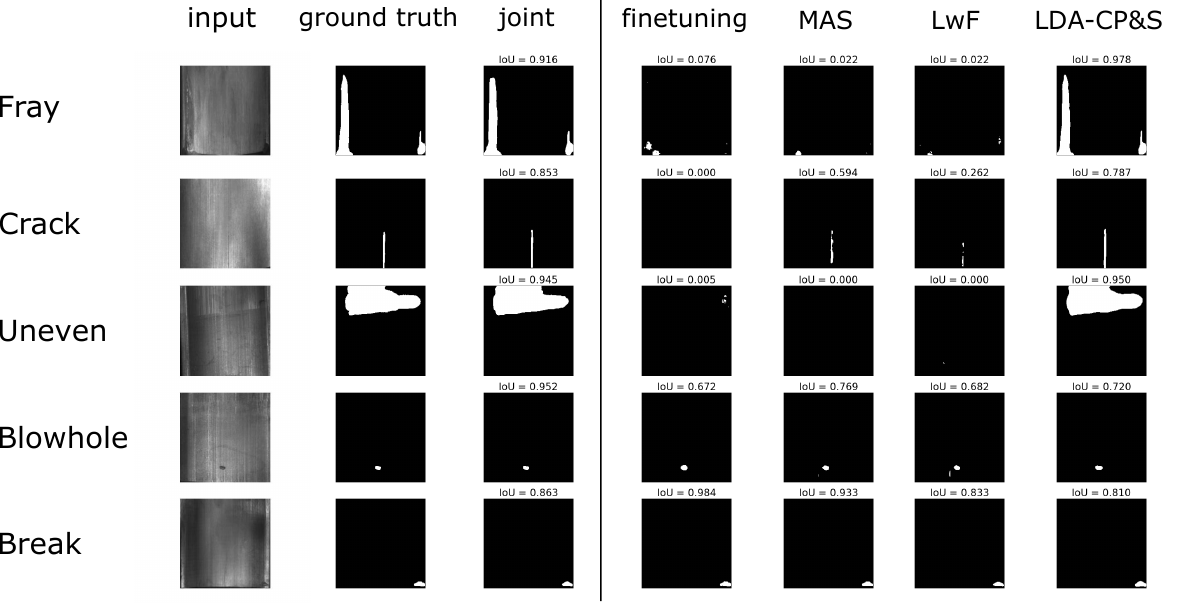}
    \caption{Visualization of defects prediction for the considered scenarios on Fray $\to$ Crack $\to$ Uneven $\to$ Blowhole $\to$ Break task ordering.}
    \label{fig:def-pred_visual}
\end{figure}

Figure \ref{fig:def-pred_visual} illustrates the model output in every learning scenario. We observe that regularization-based methods and finetuning cannot capture the defects of the first tasks in the sequence, while our LDA-CP\&S finds defects' segments close to the joint training.

\subsection{Hyperparameters choice}\label{sec:hyperparameters}

The choice of hyperparameters for pruning has a significant impact on subnetwork sparsity and, as a result, performance. In this subsection, we compare different options for the pruning hyperparameter $\alpha$ and the number of pruning iterations. A lower number of $\alpha$ and a higher number of pruning iterations lead to higher sparsity (more free connections to learn future tasks) but may cause lower segmentation performance. Also, the values for hyperparameters depend on the length of task sequences. In our work, we pre-define these hyperparameters at the beginning and do not change them during the training process.

\begin{figure*}
    \centering
    \includegraphics[width=\textwidth]{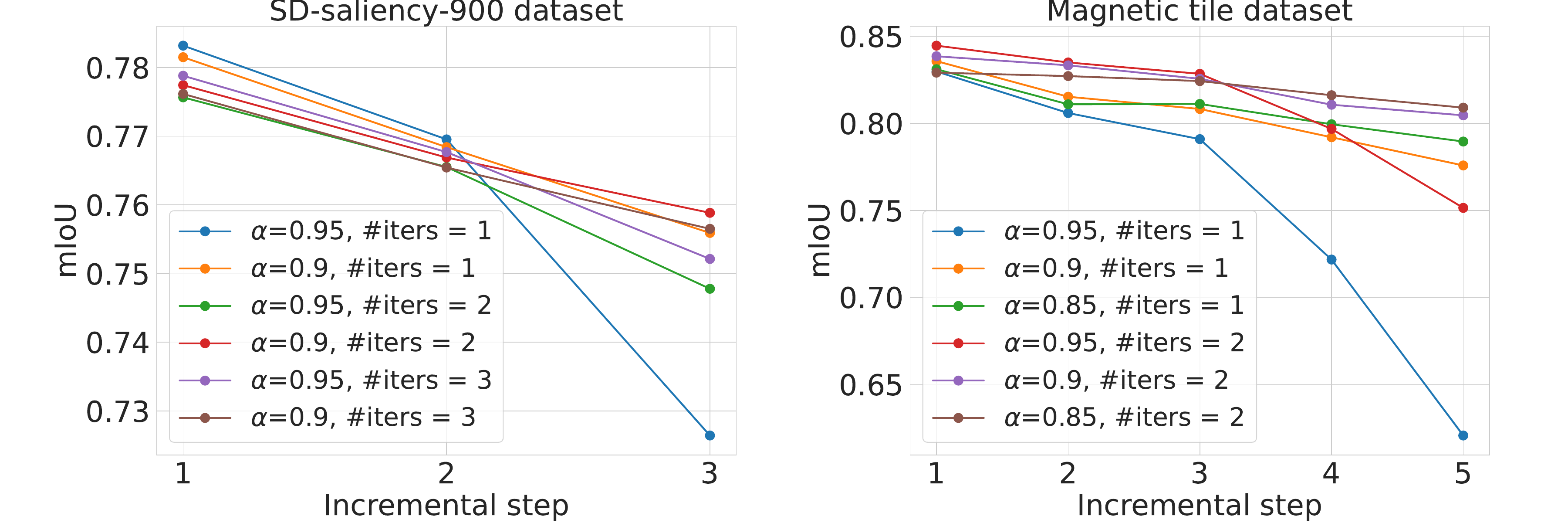}
    \caption{Comparison between different pairs of hyperparameters for the pruning step in our LDA-CP\&S method on SD-saliency-900 dataset (left) and Magnetic tile dataset (right). The results after each incremental step averaged over the number of considered task orderings.}
    \label{fig:hyperparameters_comparison}
\end{figure*}

Figure \ref{fig:hyperparameters_comparison} illustrates how different pairs of hyperparameters affect the training process for our approach. For both datasets, we clearly see that the network starts to saturate if pruning is not aggressive enough (e.g., $\alpha=0.95$ where most of the signal is conserved) because the network does not have enough free parameters for new tasks. In the case of the SD-saliency-900 dataset, we can also observe the trade-off between sparsity and mIoU score: with $\alpha = 0.9$ it is clear that pruning the network twice leads to better performance than doing it three times, as the subnetwork that results is less expressive (has fewer parameters). The results on the Magnetic tile dataset show the trade-off between learning the first tasks and the last ones: if we prune the network twice, $\alpha=0.9$ leads to better performance if there are no more than three tasks, while $\alpha=0.85$ is better suitable for longer task sequences.

\section{Conclusion}\label{sec:conclusion}

We believe smart monitoring systems should quickly adapt to new tasks without a dramatic drop in performance on previously learned ones. However, this is not the case based on the current state of the literature on surface defect inspection. Thus, there is a need for continual learning of deep neural networks for automatic surface defect segmentation such that product quality assessment is improved. By training deep learning models incrementally, we show that we can accumulate all the learned information without retraining when a new task comes to the network. In addition, we do not need to store data for retraining, which can be either not allowed or not possible due to the (lack of) availability of old datasets.

The LDA-CP\&S method that we propose successfully learns to segment the defects incrementally, without any forgetting, using only the data that is given at the current time step. Meanwhile, other methods that do not use data from previous tasks fail to remember all tasks, exhibiting considerable forgetting in segmenting previously seen defects. Overall, the performance of LDA-CP\&S is more than two times higher in terms of mean Intersection over Union score for the two datasets considered herein when compared to other continual learning methods. Moreover, it is comparable with joint training where the model has access to all the data observed up to the current incremental step. 

\backmatter



\bibliography{sn-bibliography}

\end{document}